%% file: main.tex
\documentclass[10pt,twocolumn,letterpaper]{article}

\input{includes}

\cvprfinalcopy 
\def\cvprPaperID{2139} 

\ifcvprfinal\fi
\begin{document}

\title{\textsc{NetTailor}: Tuning the architecture, not just the weights}

\author{Pedro Morgado \thanks{This work was partially funded by graduate fellowship SFRH/BD/109135/2015 from the Portuguese Ministry of Sciences and Education, NRI Grants IIS-1546305 and IIS-1637941, and NVIDIA GPU donations.} \qquad Nuno Vasconcelos\\
Department of Electrical and Computer Engineering\\
University of California, San Diego\\
{\tt\small \{pmaravil,nuno\}@ucsd.edu}
}

\maketitle

\begin{abstract}
    \input{0_abstract}
\end{abstract}

\begin{figure}[t!]
    \centering
    \includegraphics[width=0.9\linewidth]{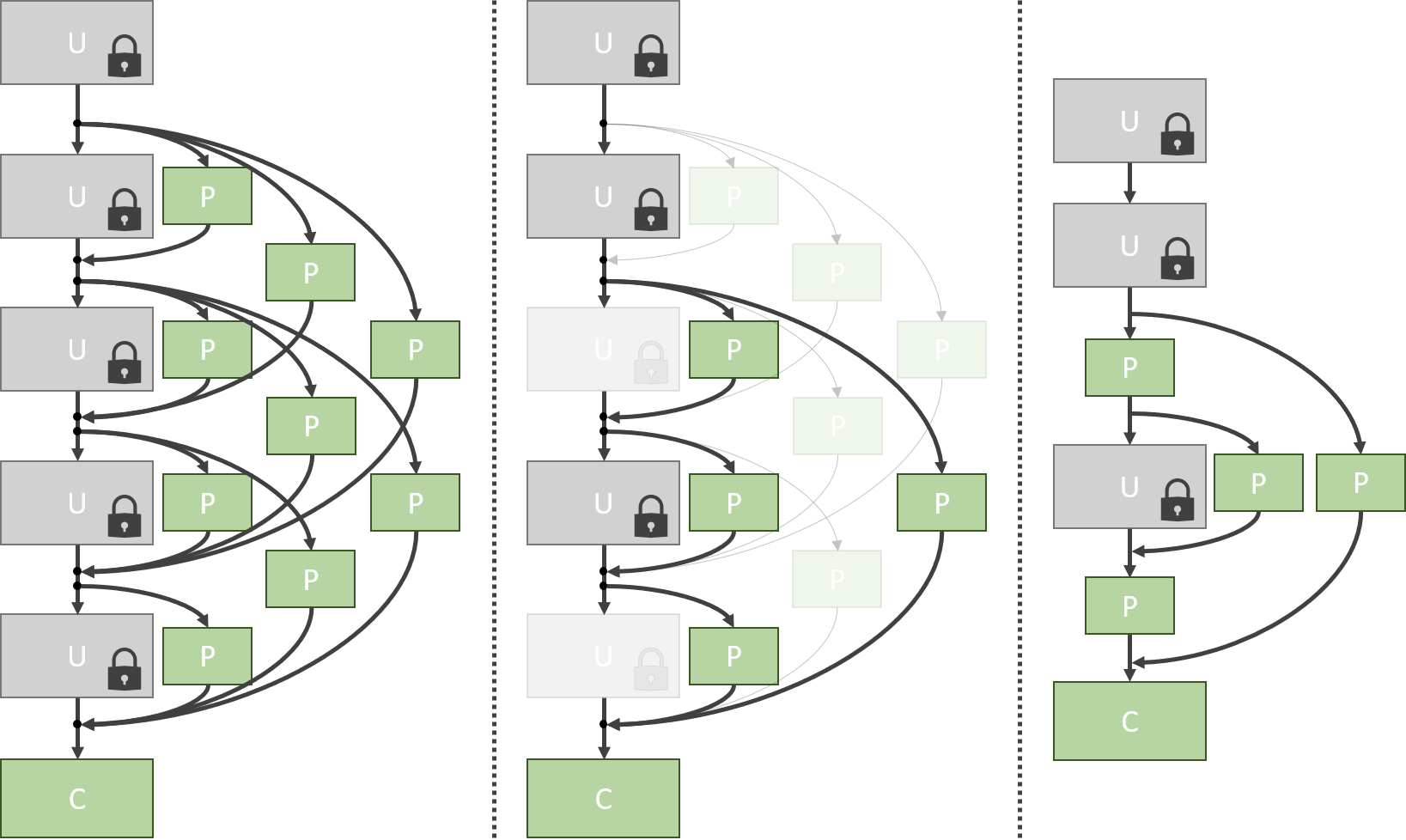}
    \caption{Architecture fine-tuning with \textsc{NetTailor}. Pre-trained blocks shown in gray, task-specific in green. Left: Pre-trained CNN augmented with low-complexity blocks that introduce skip connections.
    Center: Optimization prunes blocks of poor trade-off between complexity and impact on recognition. Right: The final network is a combination of pre-trained and task-specific blocks.}
    \label{fig:teaser}
\end{figure}

\abovedisplayskip=5pt
\belowdisplayskip=5pt

\vspace{-3pt}
\section{Introduction}
\vspace{-3pt}
\input{1_intro}

\vspace{-3pt}
\section{Related work}
\vspace{-3pt}
\input{2_related_work}

\vspace{-3pt}
\section{Method}
\vspace{-3pt}
\input{3_method}

\vspace{-3pt}
\section{Evaluation}
\vspace{-3pt}
\input{4_evaluation}

\vspace{-3pt}
\section{Conclusion}
\vspace{-3pt}
\input{5_conclusion}


{\small
\bibliographystyle{ieee}
\bibliography{refs}
}

\end{document}

%% file: includes.tex
\usepackage{cvpr}
\usepackage{times}
\usepackage{epsfig}
\usepackage{graphicx}
\usepackage{amsmath}
\usepackage{amssymb}
\usepackage{lipsum}
\usepackage{subcaption}
\usepackage{booktabs}
\usepackage[normalem]{ulem}
\usepackage[table]{xcolor}
\usepackage{multirow}
\usepackage[subtle]{savetrees}
\usepackage[font=small,skip=3pt]{caption}
\usepackage[percent]{overpic}

\definecolor{pastelblue}{RGB}{240,245,255}

\usepackage[pagebackref=true,breaklinks=true,letterpaper=true,colorlinks,bookmarks=false]{hyperref}

\newcommand{\x}{\mathbf{x}}

\definecolor{my-blue}{RGB}{51,133,198}
\definecolor{my-green}{RGB}{0,135,68}
\definecolor{my-red}{RGB}{214,57,69}



%% file: 0_abstract.tex
Real-world applications of object recognition often require the solution of \textit{multiple} tasks in a single platform.
Under the standard paradigm of network fine-tuning, an entirely new CNN is learned per task, and the final network size is independent of task complexity. 
This is wasteful, since simple tasks require smaller networks than more complex tasks, and limits the number of tasks that can be solved simultaneously.
To address these problems, we propose a transfer learning procedure, denoted \textsc{NetTailor}\footnote{\scriptsize Source code and pre-trained models available at: \\\url{https://pedro-morgado.github.io/nettailor}.}, in which layers of a pre-trained CNN are used as universal blocks that can be combined with small task-specific layers to generate new networks. 
Besides minimizing classification error, the new network is trained to mimic the internal activations of a strong unconstrained CNN, and minimize its complexity by the combination of 1) a soft-attention mechanism over blocks and 2) complexity regularization constraints. 
In this way, \textsc{NetTailor} can adapt the network architecture, not just its weights, to the target task. Experiments show that networks adapted to simple tasks, such as character or traffic sign recognition, become significantly smaller than those adapted to hard tasks, such as fine-grained recognition.
More importantly, due to the modular nature of the procedure, this reduction in network complexity is achieved without compromise of either parameter sharing across tasks, or classification accuracy.

%% file: 1_intro.tex
Real-world applications of machine learning for vision often involve the ability to solve \textit{multiple} recognition tasks.
For example, a robot should be able to decide if a door is open or closed, whether an object can be picked up or not, what is the expression on a person's face, among others.
However, attempting to design a single recognizer for all tasks is often impractical, since datasets for different tasks are not always available at the same time, and state-of-the-art models use different training procedures (e.g.~face recognition is solved through an embedding approach~\cite{schroff2015facenet}, while object recognition uses a classification loss~\cite{he2016deep}).
Instead, the standard solution is to use an off-the-shelf convolutional neural network (CNN) pre-trained on a large dataset such as ImageNet~\cite{imagenet}, MS-COCO~\cite{coco} or MIT-Places~\cite{places}, and fine-tune it to each task~\cite{yosinski2014transferable, krizhevsky2012imagenet, girshick2014rich}.

Although fine-tuning often achieve good performance on the target task~\cite{yosinski2014transferable}, this practice is quite wasteful.
First, fine-tuning produces a large network per task, {\it independently\/} of the task complexity. Hence, computing and storage requirements increase linearly with the number of tasks, with simpler tasks like optical character recognition (OCR) being as demanding as hard tasks like fine-grained recognition.
Second, although the resulting networks are derived from a common pre-trained model, they differ in {\it all\/} their parameters. Hence, as the robot switches between tasks, large arrays of parameters need to be reloaded, which may hinder operation in real-time.
Other smart computing platforms, such as consumer electronics devices, mobile devices or smart cars, also face similar problems.

In this work, we seek a solution to these problems. Layers of a large pre-trained neural network are viewed as universal blocks that can be combined to generate new networks. 
The universal blocks implement universal filters shared by all tasks. They are complemented by task-specific blocks that enable adaptation to new tasks. Given a new target task, we propose to search for the best {\it architecture\/} that combines any number of large pre-trained blocks and small task-specific blocks.
While pre-trained blocks are responsible for the bulk of feature extraction, task-specific blocks are used to 1) build the final (classification) layer, 2) simplify or even replace pre-trained blocks when possible, or 3) adjust network activations to compensate for domain differences between the source and target tasks.

Evidence for the feasibility of this idea was recently provided in~\cite{rebuffi2017learning}, where a pre-trained network is successfully adapted to multiple tasks without changing its parameters by \textit{adding} a small number of residual adaptation layers. 
In this work, however, instead of merely adding layers, we seek to {\it adapt the network architecture\/}, to tailor the network to the complexity of the new task. 
Because the process is analogous to a tailor that adjusts a pre-made suit to fit a new customer, we denote the procedure \textsc{NetTailor}. 
The main idea is illustrated in Fig.~\ref{fig:teaser}. First, we augment a pre-trained CNN with low-complexity blocks that introduce skip connections throughout the network, and a soft-attention mechanism that controls the selection of which blocks to use.
Then, we train the augmented CNN with a loss that penalizes both classification error and complexity. 
The complexity penalty favors the small task-specific blocks over the large pre-trained ones, encouraging the minimum amount of computation required by the target task.
Good classification performance is promoted with a combination of the cross-entropy loss and a variant of model distillation~\cite{hinton2015distilling}, which encourages the simplified CNN to match the performance of a classically fine-tuned CNN.
This optimization eliminates blocks with a poor trade-off between complexity and impact on recognition performance.

In sum, \textsc{NetTailor} seeks an architecture that matches the performance of standard fine-tuning, but that is as small as possible and mostly composed of universal blocks shared by many tasks. This procedure has three important properties. 
First, it enables the deployment of networks of different complexity for different tasks. 
For example, in simpler recognition problems such as digit recognition (SVHN dataset), \textsc{NetTailor} removed 73.4\% parameters, while in high-level tasks such as the recognition of everyday objects (Pascal VOC dataset) only 36.1\% of the parameters are removed.
Second, because the majority of the parameters required per task belong to shared pre-trained blocks, \textsc{NetTailor} solves more tasks with the same resources and allows task switching to be more efficient. 
On average, \textsc{NetTailor} only introduces 8\% of new task-specific parameters per task, when compared to the size of the pre-trained network.
Third, we show that pre-trained blocks can be discarded \textit{without a significant loss in performance}, achieving accuracy similar to previous transfer learning techniques.

%% file: 2_related_work.tex

\textsc{NetTailor} is related to various CNN topics.

\noindent
{\bf Transfer learning:} CNNs are routinely transferred by fine-tuning. 
\textsc{NetTailor} is a flexible transfer procedure that adjusts the network architecture (not just the weights) while keeping the majority of the parameters unchanged.

\noindent
{\bf Life-long learning \& learning without forgetting}
Intelligent systems integrate knowledge over time, leveraging what they know to solve new tasks. This ability is known as lifelong learning~\cite{thrun1995lifelong} or never-ending learning~\cite{mitchell2018never} and is usually incremental, i.e.~with tasks learned sequentially. 
Fine-tuning has two main problems for lifelong learning. First, since the original weights are modified, the number of parameters increases linearly with the number of tasks. This is wasteful since low and mid-level features can be shared across very different image domains~\cite{sharif2014cnn}. 
Second, after fine-tuning, network performance can degrade substantially on the source task \cite{goodfellow2013empirical}. 
This degradation is known as ``catastrophic forgetting" and has been the subject of various recent works, which we categorize into two groups.

The first group forces the CNN to ``remember'' the source task when training on target data~\cite{li2017LwF, kirkpatrick2017EWC,aljundi2017expert}. This is done either by 1) preventing network responses for source classes from changing significantly on images of the new task \cite{li2017LwF,aljundi2017expert}, 2) maintaining an ``episodic memory'' of images from previous tasks~\cite{lopez2017gradient,rebuffi2017icarl}, 3) preventing the reconstruction of features crucial to the source task from changing~\cite{triki2017encoder}, or 4) identifying and protecting weights critical for previous tasks~\cite{kirkpatrick2017EWC, lee2017overcoming}.
The second group retains previous task knowledge by freezing the source network and adding a small number of parameters for adaptation to the new task. 
For example, progressive neural networks~\cite{rusu2016progressive} and dynamically expandable networks~\cite{yoon2017lifelong} expand the original network by adding hidden units to each layer, and Rebuffi \etal~\cite{rebuffi2017learning, rebuffi2018efficient} add small task-specific layers, denoted residual adapters, that adapt the activations of the source network to the target task. Finally, Mallya \etal~\cite{mallya2017packnet, mallya2018piggyback} identify a small set of source weights that can be pruned or retrained to improve performance on the target task.

\textsc{NetTailor} has similarities with the second group, since it freezes pre-trained layers. Also, similarly to some methods in the first group, \textsc{NetTailor} uses source activations of intermediate layers as guidance for the activations of the new network. 
The main difference is that prior techniques do not seek to adapt the network complexity to the task requirements, which results in wasted computation when target tasks are simpler than the source task.

\noindent
{\bf Multi-task learning}
Multi-task learning (MTL) aims to improve generalization by leveraging relations between tasks~\cite{caruana1997multitask}. MTL is widely used for problems like object detection, where sharing representations between object location and classification~\cite{girshick2014rich, redmon2016you} or even segmentation~\cite{he2017mask} has led to significant gains. 
Other examples of successful MTL are head orientation and facial attribute detection~\cite{zhang2014facial, ranjan2017hyperface}, scene geometry, instance, and semantic segmentation~\cite{kendall2017multi, misra2016cross}, among others.
The main difference between MTL and transfer techniques is that MTL assumes that all tasks are performed on the same domain, usually all operating on the same image. This is not the case for transfer, where the target task belongs to a different domain, possibly very dissimilar from that of the source images.

\noindent
{\bf Domain adaptation}
Domain adaptation addresses the transfer of a task across two domains. 
When labels are available for both domains, this is usually done by fine-tuning. \textsc{NetTailor} addresses the problem that, depending on the gap between domains, there may be a need to adjust the architecture.
This is, however, different from unsupervised domain adaptation~\cite{ganin2015unsupervised, tzeng2015simultaneous,tzeng2017adversarial}, where there are no labeled data for the target domain. 
Unlike general transfer techniques like \textsc{NetTailor}, unsupervised domain adaptation is designed to bridge the gap between two datasets with \textit{exactly} the same classes, and to maximize performance on the target (unsupervised) dataset with no concern for source domain performance.

\noindent
{\bf Network compression}
Network compression aims to reduce the size of a neural network by removing weights. 
Early works~\cite{lecun1990optimal,hassibi1993second} derived near-optimal strategies to identify and remove weights of low impact on network performance. However, because these methods rely on second order derivatives of the loss function, they are impractical for deep networks.
Recently, good results have been shown with simpler procedures, such as pruning weights of low magnitude \cite{han2015learning} or introducing sparsity constraints during training \cite{zhou2016less}.
These methods reduce model size considerably but do not improve the speed of inference, due to the irregular sparsity of pruned weights. Alternative approaches advocate for ``structured sparsity'' as a means to remove entire filters~\cite{li2016pruning, molchanov2016pruning}.
\textsc{NetTailor} adopts the standard training methodology of iterative pruning (pre-train, prune, re-train), but takes the concept of structured sparsity one step further, pruning entire layers instead of weights or filters. 
However, \textsc{NetTailor} is not a network compression procedure, as layer pruning is only feasible in transfer learning, specifically when the target task is simpler than the source. 
Existing compression methods could also be used to compress the pre-trained network, further reducing the complexity of networks fine-tuned with \textsc{NetTailor}.

\noindent
{\bf Distillation}
Model distillation algorithms seek to emulate a model with a simpler, smaller or faster one. 
In \cite{buciluǎ2006model}, a strong ensemble model is used to label a large unlabeled dataset, which is then used to train a simpler model that mimics the ensemble predictions. 
Similar ideas have been used to transfer knowledge between networks with different characteristics. For example, Ba \etal~\cite{ba2014deep} demonstrate that shallow networks can mimic deeper networks while using the same amount of parameters for stronger parallelization, \cite{hinton2015distilling} and \cite{romero2014fitnets} replicate complex networks with significantly smaller or thinner ones, and \cite{chen2015net2net} transfers a previous network to a new deeper or wider network without retraining for a faster development workflow.
The student teacher paradigm used by \textsc{NetTailor} is similar to that of FitNets \cite{romero2014fitnets}, as teacher supervision is added both at the network output and internal activations. However, instead of training a new network from scratch, \textsc{NetTailor} adapts the architecture of a pre-trained network without changing most of its weights.

\noindent
{\bf Cascaded classifiers \& Adaptive inference graphs}
Cascaded classifiers~\cite{viola2001rapid}, can also significantly accelerate inference, by quickly rejecting easy negatives. 
Recent works developed these ideas within a deep learning framework, both for classification~\cite{teerapittayanon2016branchynet,huang2017multi} and detection~\cite{cai2015learning,yang2016exploit}.
By introducing early-exits, the network can classify images as soon as it reaches the desired degree of confidence~\cite{huang2017multi,teerapittayanon2016branchynet}, or anytime the decision output is expected~\cite{huang2017multi}.
Closer to \textsc{NetTailor} is the work on adaptive inference graphs (AIG)~\cite{veit2018convolutional,figurnov2017spatially}, which dynamically adjusts the network topology at test time conditioned on the image alone.
Thus, similar to \textsc{NetTailor}, both cascades and AIG methods can select which parts of the network to evaluate for each image.
However, these methods cannot effectively solve the multi-domain classification problem.
When networks are trained independently, a different network is generated per task. 
Training networks jointly requires simultaneous access to all datasets. This drastically restricts the training of networks by different developers, for different tasks, at different times, since different developers 1) may not have access to each other's data, and 2) usually lack the resources and desire to train for tasks other than their own. \textsc{NetTailor} addresses this problem by reusing a set of universal blocks shared across datasets, allowing each developer to focus on the single task of interest. It reduces both inference times and space requirements without the need for joint training on all datasets.

\noindent
{\bf Neural architecture search}
Neural architecture search (NAS) is devoted to learning new network architectures in a data-driven manner.
Typically, this is accomplished using reinforcement learning or evolutionary algorithms to update a model responsible for generating architectures so as to maximize performance~\cite{zoph2016neural,zoph2017learning}.
Since the space of possible architectures is extremely large, NAS can be quite slow and recent developments focus on accelerating the search process~\cite{liu2017progressive,liu2018darts}. 
\textsc{NetTailor} can be seen as a differentiable NAS procedure, since the network architecture is optimized for a given task. 
However, unlike general NAS, we seek a solution that reuses a set of pre-trained blocks in order to address the storage and computing inefficiencies associated with multi-domain transfer learning problems.

\noindent
{\bf Curriculum learning} Curriculum learning techniques use variations of back-propagation to improve learning effectiveness. This can be done by controlling the order in which examples are introduced~\cite{bengio2009curriculum}. Other approaches use a teacher network to enhance the learning of a student network~\cite{fan2018learning,matiisen2017teacher}. \textsc{NetTailor} uses a replica of the source network, fine-tuned on the target task, as a teacher for the learning of the simplified network.

%% file: 3_method.tex
In this section, we introduce \textsc{NetTailor}.

\subsection{Task transfer}
\vspace{-3pt}

A CNN implements a function 
\begin{equation}
    f(\x)=(G_L \circ G_{L-1} \circ \cdots \circ G_1) (\x).
    \label{eq:cnn}
\end{equation}
by composing $L$ computational blocks $G_l(\x)$ consisting of simple operations, such as convolutions, spatial pooling, normalization among others.
For object recognition, $\x$ is an image from a class $y \in \{1, \ldots, C\}$, and $f(\x)\in[0,1]^C$ models the posterior class probability $P(y|\x)$.
While the blocks $G_l(\x)$ differ with the CNN model, they are often large, both in terms of computation and storage. 
For example, under the ResNet model, each $G_l(x)$ is formed by two $3\!\times\!3$ convolutions, or in deeper versions a ``bottleneck'' block containing two $1\!\times\!1$ and one $3\!\times\!3$ convolutions~\cite{he2016deep}.

Since CNN training requires a large dataset, such as ImageNet~\cite{imagenet}, Places~\cite{places} or COCO~\cite{coco}, not available for most applications, CNNs are rarely learned from scratch. 
Instead, a CNN pre-trained on a large dataset is fine-tuned on a new task. In this case, the original task is denoted as the {\it source\/}  and the new one as the {\it target\/} task. Fine-tuning adjusts the weights of the blocks of~\eqref{eq:cnn}, while maintaining the network architecture.
Hence, independently of the complexity of the new task, the computational and storage complexity remain large. This is undesirable for target tasks simpler than the source task, especially for applications that have computational or storage constraints, such as mobile devices. 

\subsection{NetTailor}
\vspace{-3pt}

In order to avoid these problems, task transfer should ideally have two properties. 
First, rather than reusing entire networks, it should reuse network blocks. In particular, it should be possible to add or remove blocks to best adapt the architecture to the new task, not just its weights. This way, if the target task is much simpler than the source task, network size could decrease significantly. 
Second, new networks should reuse existing pre-trained blocks to the largest possible extent, in order to minimize the number of parameters to be learned. 
Reusing blocks is particularly crucial for memory constrained implementations (e.g., robotics or mobile devices), because it allows sharing of blocks across tasks. In this case, since only a fraction of (task-specific) parameters need to be switched and stored per task, both the costs of task switching and model storage remain low. 

In this work, we introduce a new transfer technique, denoted \textsc{NetTailor} that aims to achieve these goals. The \textsc{NetTailor} procedure illustrated in Fig.~\ref{fig:teaser} can be summarized as follows.
\begin{enumerate}
    \itemsep-1pt
    \item Train the {\it teacher network\/} by fine-tuning a pre-trained network on the target task.
    \item Define the {\it student network\/} by {\it augmenting\/} the pre-trained network with task-specific low-complexity {\it proxy layers\/}.
    \item Train the task-specific parameters of the student network on the target task to {\it mimic\/} the internal activations of the teacher, while imposing {\it complexity constraints\/} that encourage the use of low-complexity proxy layers over high-complexity pre-trained blocks.
    \item Prune layers with low impact on network performance.
    \item Fine-tune the remaining task-specific parameters.
\end{enumerate}
While we only experimented with teacher networks that are learned by fine-tuning (step 1), \textsc{NetTailor} could also be used with any transfer technique that produces a teacher that preserves the architecture of the pre-trained network. 
We focused on fine-tuning due to its popularity and high performance for most tasks where a reasonably sized dataset is available for training~\cite{krizhevsky2012imagenet,yosinski2014transferable}.
Layer pruning (steps 4 and 5) is performed using operations common in the network compression literature~\cite{han2015learning,zhou2016less}, and is briefly described in Section~\ref{sec:pruning}.
We now discuss steps 2 and 3 in detail.

\begin{figure}[t]
    \centering
    \includegraphics[width=0.7\linewidth]{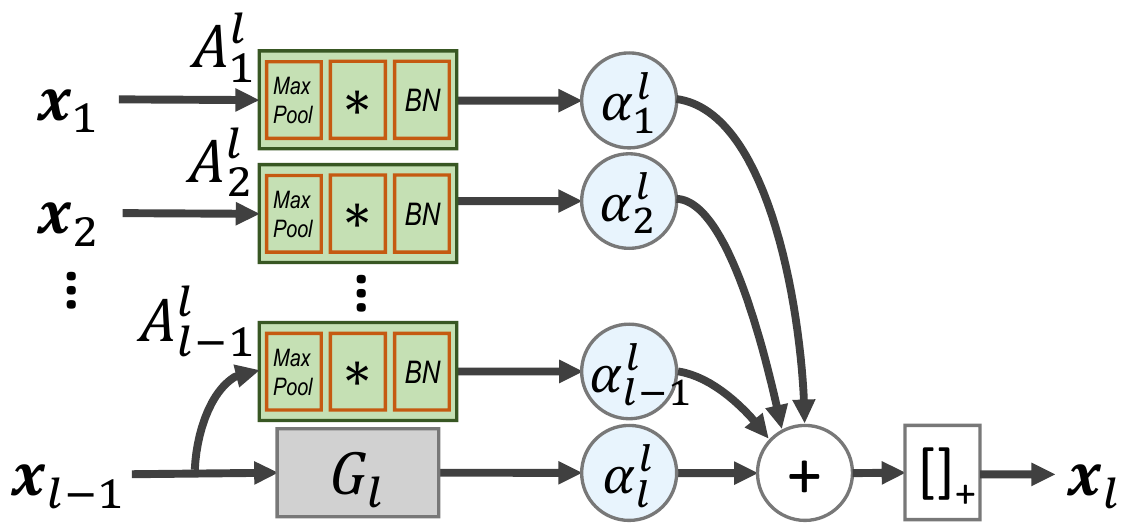}
    \caption{Augmentation of pre-trained block $G_l$ at layer $l$ with multiple proxy layers $A_p^l$. $\x_i$ represents the network activation after layer $i$.}
    \label{fig:pooling}
\end{figure}

\subsection{Architecture of the student network}
\vspace{-3pt}
The main architectural component introduced in this work is the augmentation of the pre-trained network $f(\x)=(G_L \circ G_{L-1} \circ \cdots \circ G_1) (\x)$ with the complexity-aware pooling block of Figure~\ref{fig:pooling}. 
Starting from the pre-trained model, each layer $G_l$ is augmented with a set of lean proxy layers $\lbrace A_p^l(\cdot)\rbrace_{p=1}^{l-1}$ that introduce a skip connection between layers $p$ and $l$. As the name suggests, proxy layers aim to approximate and substitute the large pre-trained blocks $G_l(\cdot)$ whenever possible.
The output activation $\x_l$ of layer $l$ is then computed by pooling the output of the $l^{th}$ pre-trained block $G_l(\cdot)$ and proxies $A_p^l(\cdot)$
\begin{equation}
    \textstyle \x_l = \alpha_l^l G_l(\x_{l-1}) + \sum_{p=1}^{l-1} \alpha_p^l A_p^l(\x_p),
    \label{eq:fwd_pooling}
\end{equation}
where $\lbrace \alpha_p^l\rbrace_{p=1}^{l}\in[0,1]$ are a set of scalars that enable or disable the different network paths.

\begin{figure*}[t!]
    \centering
    \begin{tabular}{c|c|c}
        \rowcolor{white}
        \begin{overpic}[height=2.3cm]{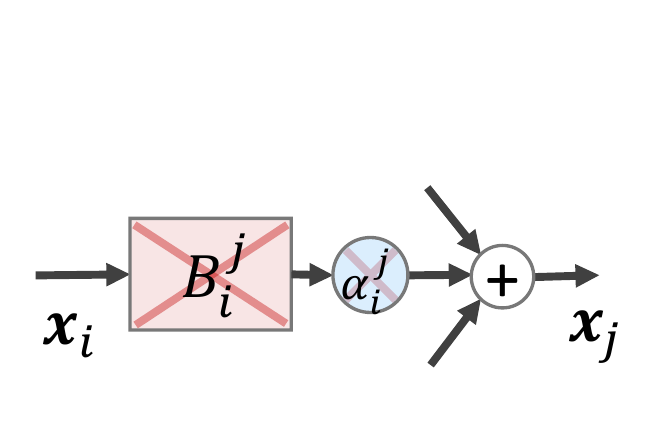}
            \put (50,00) {(a)}
        \end{overpic}
        & 
        \begin{overpic}[height=2.3cm]{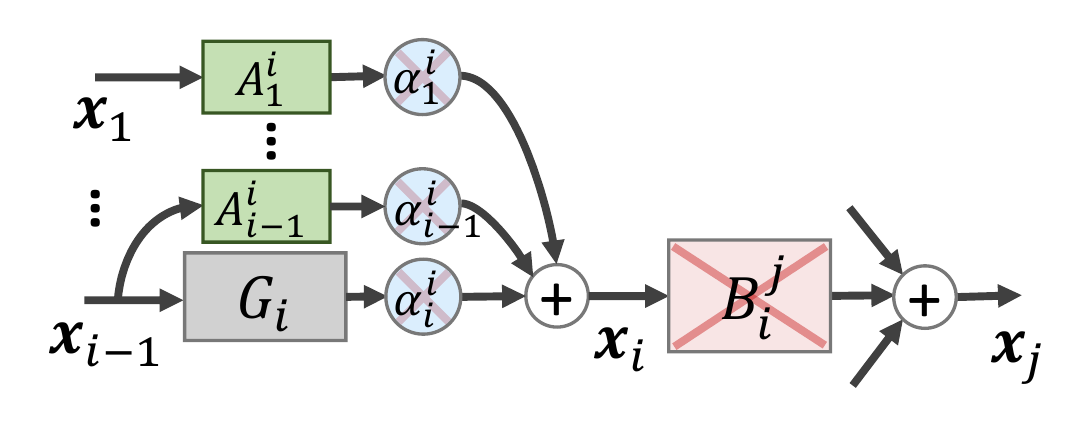}
            \put (40,00) {(b)}
        \end{overpic} 
        &
        \begin{overpic}[height=2.3cm]{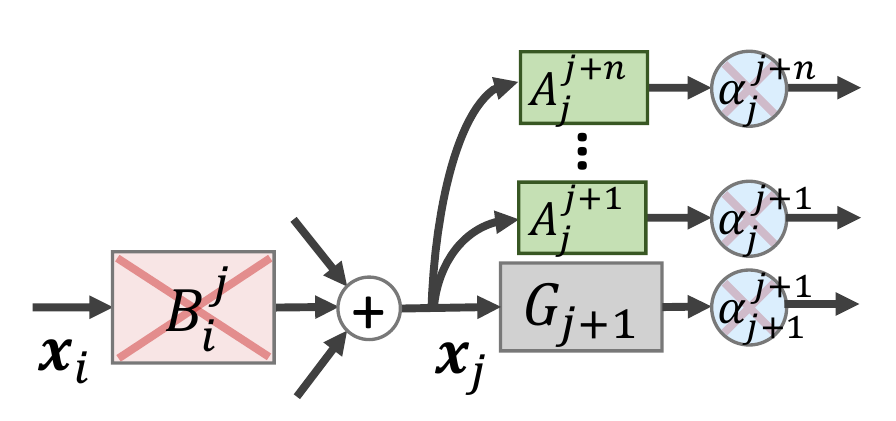}
            \put (60,00) {(c)}
        \end{overpic} 
    \end{tabular}
    \caption{Block removal criteria. (a) Self-exclusion. (b) Input exclusion. (c) Output exclusion.}
    \label{fig:exclusion}
    \vspace{5pt}
\end{figure*}

Two steps are taken to reduce the number of task-specific parameters. 
The first is to use proxy layers of low-complexity. 
Specifically, $A_p^l(\cdot)$ is composed of 1) a spatial max-pooling block that converts activations from the spatial resolution of $\x_p$ into that of $\x_l$, and 2) a $1\!\times\!1$ convolution (with batch normalization) that projects the input feature map $\x_p$ into the desired number of channels for $\x_l$.
Thus, in comparison to the standard ResNet block which contains two $3\!\times\!3$ convolutions, each proxy $A_p^l(\cdot)$ contains only $\frac{1}{18}$ of the parameters and performs only $\frac{1}{18}$ of floating point operations.
Second, proxy layers are forced to compete with each other to minimize the propagation of redundant information through the network.
This is accomplished by introducing a set of auxiliary parameters $a_p^l$ and computing $\alpha_p^l$ as the softmax across all paths merging into layer $l$
\begin{equation}
    \textstyle \alpha_p^l = \frac{e^{a_p^l}}{\sum_k e^{a_k^l}}.
    \label{eq:alphas}
\end{equation}

Finally, while the description above implies a dense set of low-complexity proxies, connecting the outputs of all layers $i<l$ to that of layer $l$, we found this to be often unnecessary (see Section~\ref{sec:analysis}). Therefore, we limit the number of proxies in~\eqref{eq:fwd_pooling} to the closest $k$, and use
\begin{equation}
    \textstyle \x_l = \alpha_l^l G_l(\x_{l-1}) + \sum_{p=\max(l-k,1)}^{l-1} \alpha_p^l A_p^l(\x_p).
    \label{eq:fwd_pooling_skip}
\end{equation}
Figure~\ref{fig:teaser} illustrates the initial student architecture for $k=3$.

\subsection{Tailoring the student to the target task}
\vspace{-3pt}


The student network seeks a trade-off of two goals: low complexity and performance similar to the teacher. 

\vspace{-7pt}
\subsubsection{Constraining student complexity}
\vspace{-3pt}
\label{sec:complexity}

In the complexity-aware pooling block of Fig.~\ref{fig:pooling}, scalars $\lbrace \alpha_p^l\rbrace_{p=1}^{l}$ act as a soft-attention mechanism that selects which blocks to use for the target task.
Let $B_i^j(\cdot)$ denote the computational block associated with path $i\rightarrow{}j$, i.e.~$B_i^j(\cdot)\!=\!G_i(\cdot)$ if $i\!=\!j$ or $B_i^j(\cdot)\!=\!A_i^j(\cdot)$ otherwise. Then, block $B_i^j(\cdot)$ can be removed if one of three conditions hold:
\begin{itemize}
    \itemsep-1pt
    \item Self-exclusion (Fig.~\ref{fig:exclusion}a): path $i\rightarrow{}j$ is excluded, i.e.~$\alpha_i^j=0$;
    \item Input exclusion (Fig.~\ref{fig:exclusion}b): all paths merging into node $i$ are excluded, i.e.~${\alpha_k^i=0}, \forall k\leq i$;
    \item Output exclusion (Fig.~\ref{fig:exclusion}c): all paths departing from node $j$ are excluded, i.e.~${\alpha_j^k=0}, \forall k>j$ and ${\alpha_{j+1}^{j+1}=0}$.
\end{itemize}
Note that while self-exclusion only allows the removal of a single block, both input and output exclusion remove multiple blocks simultaneously. 
For example, if all paths merging into node $i$ are excluded, then all blocks departing from this node have no viable input and can be removed. 
Similarly, if all paths departing from node $j$ are excluded, then all blocks merging into this node will end up being ignored and can be removed as well.

To tailor the architecture to the target task, the set of scalars $\lbrace \alpha_p^l\rbrace_{p=1}^{l}$ should enable high performance, but minimize the expected network complexity.
Let $R_{self}^{i,j}$, $R_{inp}^{i}$ and $R_{out}^{j}$ denote the events associated with conditions 1, 2 and 3, respectively, and $\mathcal{C}_i^j$ the complexity of block $B_i^j$. 
Then, the expected complexity of block $B_i^j$ is
\begin{equation} 
    E\left[\mathcal{C}_i^j\right] =
    \mathcal{C}_i^j \left(1-P(R_{self}^{i,j} \cup R_{inp}^{i} \cup R_{out}^{j})\right).
     \label{eq:block_comp}
\end{equation}
Under the assumptions that events $R_{self}^{i,j}$, $R_{inp}^{i}$ and $R_{out}^{j}$ are disjoint, and events $R_{self}^{i,k}$ are all independent, the probability of (\ref{eq:block_comp}) is given by
\begin{equation}
    \small
    P(R_{self}^{i,j} \cup R_{inp}^{i} \cup R_{out}^{j}) = P(R_{self}^{i,j}) + P(R_{inp}^{i}) + P(R_{out}^{j})
    \label{eq:disjointprob}
\end{equation}
with
\begin{align}
    \textstyle P(R_{inp}^{i})  & =\textstyle P\left(\cap_{k\leq i} R_{self}^{k,i}\right) = \prod_{k\leq i} P(R_{self}^{k,i}) \label{eq:input}
\end{align}
\begin{align}
    \textstyle P(R_{out}^{j})  & =\textstyle P\left(\cap_{k>j} R_{self}^{j,k} \cap R_{self}^{j+1,j+1}\right) \nonumber \\
    &=\textstyle P(R_{self}^{j+1,j+1}) \cdot \prod_{k>j} P(R_{self}^{j,k}).  \label{eq:output}
\end{align}

Finally, by modeling the probability of self-exclusion by $P(R_{self}^{i,j}) = r_i^j = 1-\alpha_i^j$, then \eqref{eq:block_comp} becomes
\begin{align} 
    \textstyle E\left[\mathcal{C}_i^j\right] =
    \mathcal{C}_i^j \left(1 - r_i^j - \prod_{k\leq i} r_k^i - r_{j+1}^{j+1}\prod_{k>j} r_j^k\right),
     \label{eq:block_comp_lin}
\end{align}
and the expected network complexity
\begin{align}
    \textstyle E[\mathcal{C}] = \sum_{i,j} E\left[\mathcal{C}_i^j\right].
    \label{eq:net_complexity}
\end{align}
Although the exclusion events may not be disjoint or independent, the minimization of \eqref{eq:net_complexity} still provides the desired incentive towards the use of low-complexity proxies. Hence, we use \eqref{eq:net_complexity} as a differentiable complexity penalty explicitly enforced during training.

\vspace{-10pt}
\subsubsection{Mimicking the teacher}
\vspace{-3pt}
\label{sec:proxy}

The teacher network is obtained by fine-tuning a pre-trained network for the target task.
To transfer this knowledge to the student network, the latter is encouraged to match the internal activations of the teacher, by adding an $L_2$ regularizer
\begin{align}
    \textstyle \Omega = \sum_{l} \|\x_l^t-\x_l\|^2,
    \label{eq:proxy_loss}
\end{align}
where $\x_l^t$ is the activation of $l^{th}$ block of the teacher network, $\x_l$ the corresponding activation of the student network given by~\eqref{eq:fwd_pooling}, and the sum is carried over all internal blocks as well as network outputs (prior to the softmax).

\vspace{-10pt}
\subsubsection{Loss function}
\vspace{-3pt}
\textsc{NetTailor} optimizes all task-specific parameters of the student network end-to-end to meet three goals: 1) minimize classification loss on the target task, 2) minimize network complexity and 3) minimize the approximation error to the teacher network. 
Given a target dataset $\mathcal{D}=\{\x_i, y_i\}$ of images $\x_i$ and labels $y_i$, this is accomplished by minimizing the loss function 
\begin{align}
    \textstyle L = \sum_i L_{cls}(f(\x_i), y_i) + 
    \gamma_1 E[\mathcal{C}] +
    \gamma_2 \Omega,
    \label{eq:loss}
\end{align}
where $f(\cdot)$ denotes the output of the student network, $L_{cls}(f(\x), y)$ is the cross-entropy loss between the network prediction $f(\x)$ and ground-truth label $y$, $E[\mathcal{C}]$ is the expected network complexity of \eqref{eq:block_comp_lin} and \eqref{eq:net_complexity}, $\Omega$ is the teacher approximation loss of~\eqref{eq:proxy_loss}, and $\gamma_1$ and $\gamma_2$ two hyper-parameters that control the importance of each term.

\begin{figure*}[t!]
  \centering
  \includegraphics[width=0.92\linewidth]{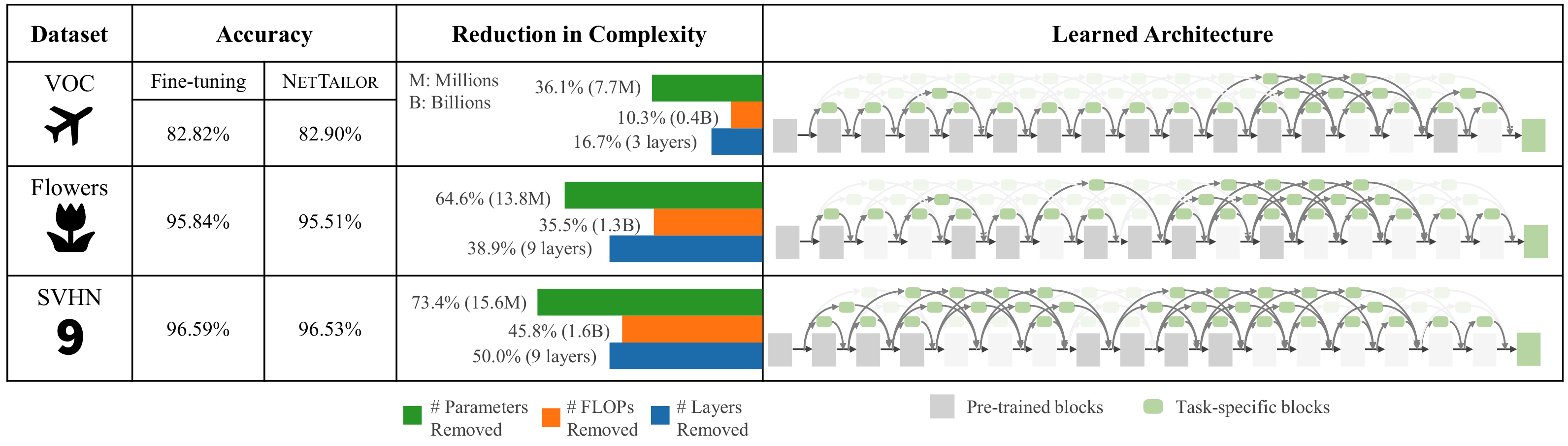}
  \caption{Reduction of network complexity and final architecture after adapting ResNet34 to three datasets using \textsc{NetTailor}.}
  \label{fig:dataset}
\end{figure*}

\subsection{Pruning and fine-tuning}
\vspace{-3pt}
\label{sec:pruning}
After training the student network, the magnitude of the scalars $\alpha_i^j$ reflects the importance of each block, with values close to zero indicating a low impact on network performance.
Given this observation, we threshold the scalars $\alpha_i^j$, and use the three exclusion conditions outlined above to remove all unnecessary blocks.
In order to enable better control over the trade-off between performance and complexity, proxies and pre-trained blocks are removed using different pruning schemes.
Since proxy layers are both small and crucial for the adaptation to the target task, we define a very low threshold $\theta$ (typically 0.05) and only remove proxies with $\alpha_i^j<\theta$.
As for pre-trained layers, we first rank their importance by the values of $\alpha_i^i$, and remove the $k$ least important blocks.
Finally, in order to recover from the removal of network components, all remaining task-specific layers are fine-tuned to minimize the loss of~\eqref{eq:loss} without complexity constraints ($\gamma_1=0$).

%% file: 4_evaluation.tex
We conducted a series of experiments to evaluate the \textsc{NetTailor} procedure.
Section~\ref{sec:analysis} provides an in-depth analysis of the impact of important variables such as the complexity of the target task, the depth of the pre-trained network, the importance of the teacher and the number of skips in the student network.
Then, to demonstrate the effectiveness of the proposed procedure, Section~\ref{sec:sota} compares \textsc{Net-Tailor} to prior work on several datasets.

\subsection{Analysis}
\vspace{-3pt}
\label{sec:analysis}
We analyze \textsc{Net-Tailor} using three classification datasets of varying characteristics: SVHN, VGG-Flowers and Pascal VOC 2012. SVHN~\cite{svhn} is a large digit recognition dataset containing 100k images of street view house numbers. VGG-Flowers~\cite{flowers} is a small fine-grained dataset composed by 8k images distributed across 102 flower species. PASCAL VOC 2012~\cite{voc12} is a dataset for the detection of a small number (20) of common objects. While VOC was designed for object detection, we test our method on the classification task alone. We used ground-truth bounding boxes to crop all objects with a 20\,\% margin and re-sampled the dataset to avoid large class imbalances.
We used standard training and test sets in all cases.

\noindent
{\bf Training details}
We now describe the standard implementation of \textsc{NetTailor} which, unless otherwise specified, is used throughout our experiments.
Global blocks are obtained by pre-training a large CNN model on ImageNet (ResNet34 in most of our experiments) and remain unchanged afterward in order to share them across tasks.
For each target task, the teacher is trained by fine-tuning the pre-trained network.
The student is assembled by augmenting the pre-trained blocks with three skip connections per layer, and all task-specific parameters (i.e.~final classifier, proxy layers and scalars $\alpha$) are trained to minimize the loss of~\eqref{eq:loss} with $\gamma_1=0.3$ (complexity constraints) and $\gamma_2=10$ (teacher loss).
In the complexity constraints of \eqref{eq:block_comp_lin}, the complexity $\mathbb{C}_i^j$ is defined as the number of FLOPs of each block normalized by the total number of FLOPs of the pre-trained network. This definition makes pre-trained layers about 20 times more expensive than proxy layers.
One critical detail is the initialization of the scalars $\alpha$ to initially favor pre-trained over task-specific blocks. This initialization provided a good starting point for learning (i.e.~similar to the pre-trained network alone) and reduced overfitting.
Specifically, we set the initial value of $a_i^i$ to $2$ for all $i$ (i.e.~pre-trained blocks), and $a_i^j$ to $-2$ for all $i\neq j$ (i.e.~proxies).
After training the student network, we remove all proxies with $\alpha_i^j<0.05$ and the $k$ least important pre-trained blocks (as ranked by the values of $\alpha_i^i$).
Finally, we fine-tune the remaining task-specific parameters to minimize the loss of~\eqref{eq:loss} without complexity constraints $\gamma_1=0$.
The pruning and retraining steps are repeated multiple times with different values of $k$, and the leanest model that achieves a target accuracy within 0.5\% of the teacher network is chosen as the final architecture.
All hyper-parameter values were chosen based on early experiments and used for all three datasets, as they tend to provide a good trade-off between accuracy and network complexity. 
A study of some of these parameters is provided below.
As usual with classification problems, we used stochastic gradient descent with momentum in all training steps.

\begin{figure}[t!]
  \centering
  \includegraphics[width=0.92\linewidth]{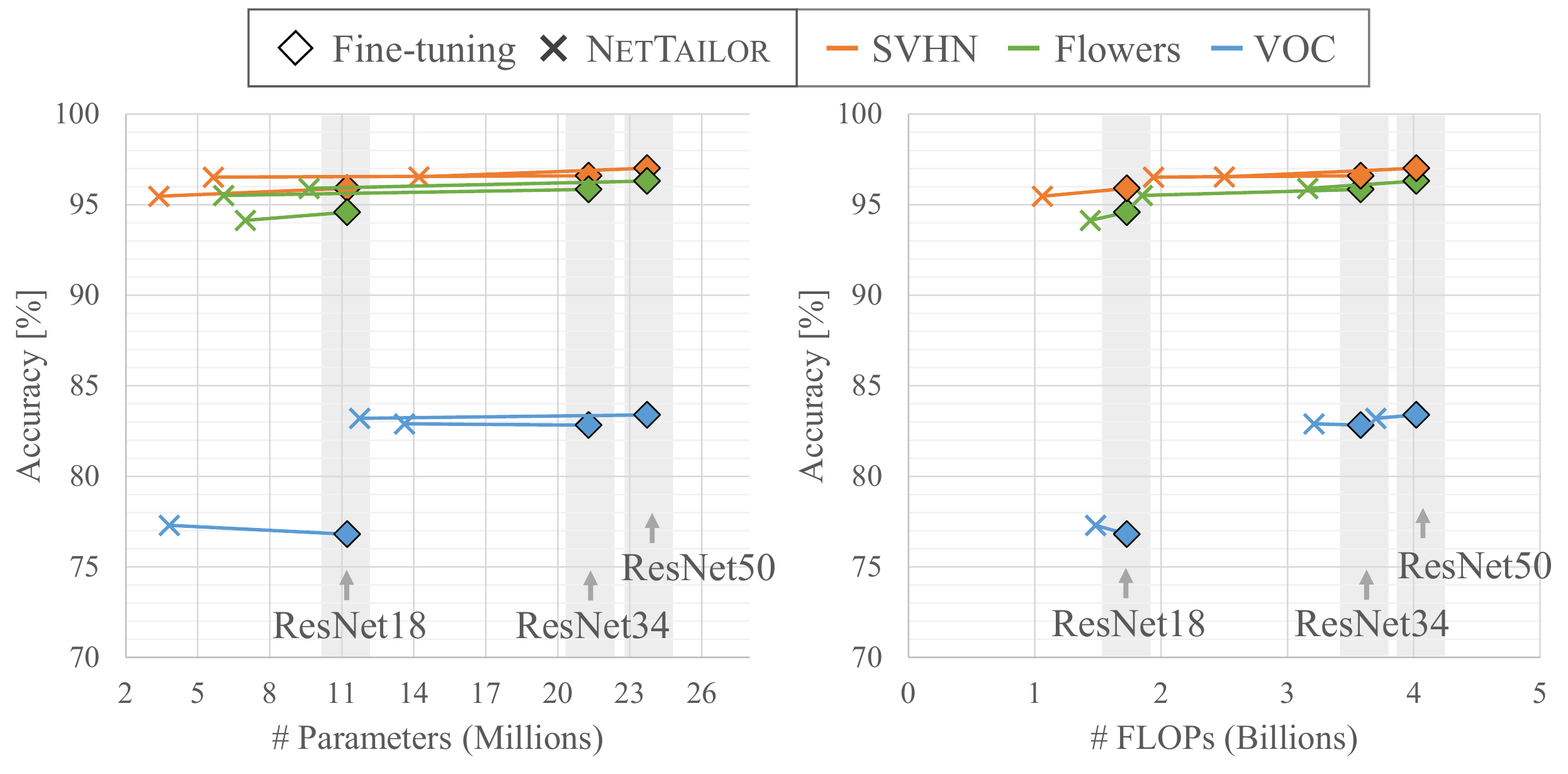}
  \caption{Accuracy vs.~complexity of models of increasing depth. Diamonds represent the fine-tuned model and crosses the model obtained with \textsc{NetTailor}. The lines connect fine-tuned models to their adapted counterparts.}
  \label{fig:model_size}
\end{figure}

\begin{figure}[t!]
  \centering
  \includegraphics[width=0.92\linewidth]{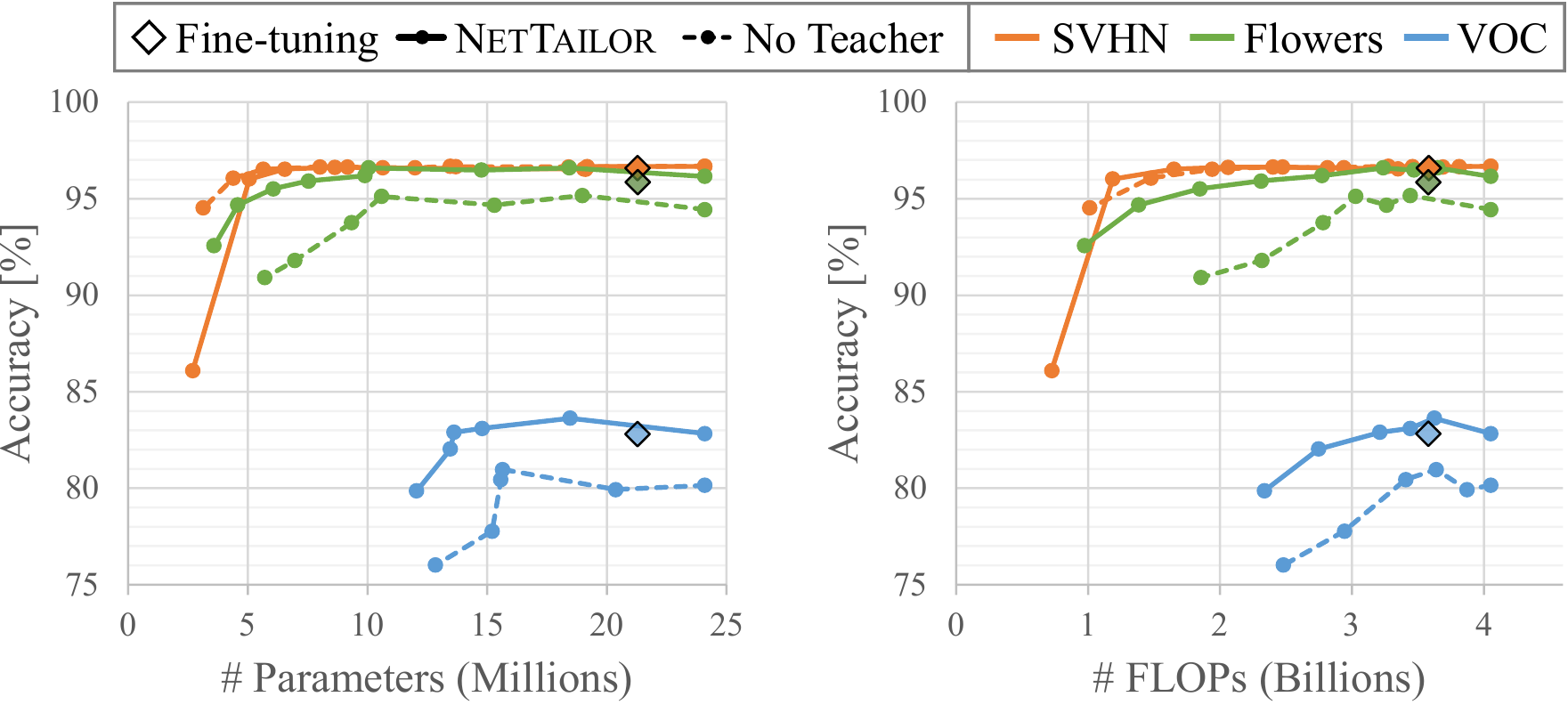}
  \caption{Accuracy vs.~complexity of models discovered by \textsc{NetTailor} with and without the teacher. Right-most dots represent unpruned networks and subsequent ones networks with increasing numbers of removed layers.}
  \label{fig:teacher}
\end{figure}

\noindent
\textbf{Effectiveness of \textsc{NetTailor} on various datasets:}
To study the impact of dataset complexity, we tuned the ResNet34 architecture using \textsc{NetTailor} and measured the maximum achievable reduction in network complexity that retains performance similar to fine-tuning. 
The results are shown in Fig.~\ref{fig:dataset} for three different datasets. We list the total number of layers, parameters (global and task-specific) and FLOPs removed from the pre-trained network by \textsc{NetTailor}. We also display the final learned architecture for each task.
Fig.~\ref{fig:dataset} shows that networks trained for simpler tasks, such as SVHN, are the most heavily pruned, with 9 out of 18 pre-trained blocks removed. This results in a drastic 73.4\,\% reduction in total parameters and a 45.8\,\% reduction in FLOPs.
For simpler tasks, most residual blocks are unnecessary and fine-tuning likely converts them into transformations close to the identity, which can be replaced by low complexity proxies.
\textsc{NetTailor} also obtains significant reductions for the more complex Flowers and VOC datasets.
Overall, the results of Fig.~\ref{fig:dataset} show that, for many applications, large pre-trained networks can be significantly reduced, both in size and speed, without loss of performance. Furthermore, because the pre-trained blocks remain unchanged, only a small number of new parameters is introduced per task: 1.90M (million) for VOC, 1.88M for flowers and 1.85M for SVHN (pre-trained ResNet34 blocks have a total of 21.29M parameters).

\noindent
\textbf{Depth of pre-trained model:}
To understand the effectiveness of \textsc{NetTailor} when applied to networks of increasing depth, we used the ResNet model family: ResNet18, ResNet34, and ResNet50. 
Unlike ResNet18 and ResNet34, ResNet50 blocks have a bottleneck architecture, where input maps are projected into a low-dimensional space by a $1\!\times\!1$ convolution, then processed by a $3\!\times\!3$ convolution, projected back into the high-dimensional space through a $1\!\times\!1$ convolution, and added to the residual link.
Due to the high-dimensionality of the output, this block architecture does not allow the use of proxies as defined in Fig.~\ref{fig:pooling}, since $1\!\times\!1$ convolutions in the high-dimensional space are still expensive.
Instead, to keep the complexity of each proxy at about $\frac{1}{20}$ of the pre-trained block, we employ a bottleneck structure to the proxy as well, i.e.~we employ two consecutive $1\!\times\!1$ convolutions with batch-norm. The first projects the input into a low-dimension space (4 times smaller than the bottleneck of the pre-trained block), and the second restores the input dimensionality.

The results depicted in Fig.~\ref{fig:model_size} show that \textsc{NetTailor} can produce architectures that achieve the performance of a larger CNN (e.g.~ResNet50) with the same or fewer parameters as a smaller one (ResNet18). This is especially important for more complex problems, where network depth has a bigger impact on performance. For example, for the VOC dataset, \textsc{NetTailor} is able to reduce ResNet50 to only 11.7M parameters, only 0.5M more than ResNet18, but with much higher performance (83.2\,\% vs 79.6\,\% accuracy). Reduction in inference speed, however, was not as drastic for the VOC dataset, since \textsc{NetTailor} mostly removed high-level layers which contain most of the parameters but only account for a small number of operations.

\input{table-skips.tex}
\input{table-piggyback.tex}
\input{table-decathlon.tex}

\noindent
\textbf{Teacher supervision:}
Fig.~\ref{fig:teacher} shows the advantage of using a fine-tuned network as the teacher.
Each line in Fig.~\ref{fig:teacher} shows the performance achieved by the model after removing different numbers of blocks $k$ (smaller values of $k$ produce models of higher complexity).
As can be seen, removing the teacher leads to significant loss in performance, regardless of the number of removed blocks, with the student network never achieving the same performance as fine-tuning.
The exception to this trend is the SVHN dataset, which is a simple dataset with a large number of images. This indicates that the skip architecture of Fig.~\ref{fig:teaser} is prone to overfitting in smaller datasets, but teacher supervision provides an effective solution to this problem.

\noindent
\textbf{Student architecture:}
We also compare different student architectures, by augmenting ResNet34 with 1, 3, 5 or a dense set of skip proxies.
The results presented in Table~\ref{tab:skips} show that augmenting the student architecture with a dense set of proxies can be beneficial for simpler datasets like SVHN. This is because accurate digit classification depends largely on lower-level features that are directly bypassed into the classification layer by proxies that skip a large number of blocks.
By contrast, dense skips are unnecessary for harder datasets, such as Flowers or VOC, with \textsc{NetTailor} removing most of the long reach proxies.
Also, as shown in Table~\ref{tab:skips}, directly imposing a limit on the number of proxies per layer leads to more significant reductions in complexity for the same performance level.

\subsection{Comparison to prior work}
\vspace{-3pt}
\label{sec:sota}
Finally, we compare \textsc{NetTailor} to prior transfer learning methods designed for the efficient classification of multiple domains.
We follow two experimental protocols. The first protocol described in~\cite{mallya2018piggyback} consists of five datasets: CUB~\cite{cub}, Stanford Cars~\cite{cars}, Oxford Flowers~\cite{flowers}, WikiArt~\cite{wikiart} and Sketch~\cite{sketch}. Following~\cite{mallya2018piggyback}, we use the same train/test set splits, and apply the \textsc{NetTailor} procedure to the same backbone network, ResNet50, with an input size 224x224. 
The second protocol is the visual decathlon benchmark~\cite{rebuffi2017learning} and consists of ten different datasets including ImageNet, Omniglot, German Traffic Signs, among others.
We use the same train/validation/test sets provided in~\cite{rebuffi2017learning} which contain images resized to a common resolution of 72 pixels.
Similar to~\cite{rebuffi2018efficient}, we also use a wide residual network~\cite{zagoruyko2016wide} with 26 layers pre-trained on ImageNet.
Results are reported using both top-1 accuracy and the ``decathlon score''~\cite{rebuffi2017learning} which pools all results in a single metric that accounts for the different difficulty of each task.

Table~\ref{tab:piggyback} compares the results of \textsc{NetTailor} to several methods. Feature extraction computes features from a pre-trained network, which are then used to build a simple classifier. While feature extraction shares most weights across datasets, differences between the source and target domains cannot be corrected, thus achieving low performance. 
More refined methods, such as PackNet~\cite{mallya2017packnet} and Piggyback~\cite{mallya2018piggyback}, try to selectively adjust the network weights in order to remember previous tasks, or freeze the backbone network and learn a small set of task-specific parameters (a set of masking weights in the case of Piggyback) that is used to bridge the gap between source and target tasks.
All these methods ignore the fact that source and target datasets can differ in terms of difficulty, and thus the \textit{architecture} itself should be adjusted to the target task, not just the weights. 
As seen in Table~\ref{tab:piggyback}, these methods are not competitive with \textsc{NetTailor}, which can significantly reduce the network complexity both in terms of model size and inference speed. 
\textsc{NetTailor} outperforms all approaches in all datasets, improving the classification accuracy of the second best method in four out of five datasets, while requiring an average of $46\%$ fewer parameters and $22\%$ fewer FLOPS.

Comparisons in the Visual Decathlon benchmark show similar findings. 
In addition to Piggyback~\cite{mallya2018piggyback}, we also compared to Learning without Forgetting (LwF)~\cite{li2017LwF}, deep adaptation networks (DAN)~\cite{rosenfeld2018incremental} and parallel Residual Adapters (ResAdapt)~\cite{rebuffi2018efficient}. 
LwF learns a new network per task that retains the responses of the original ImageNet model. 
Hence, similar to fine-tuning, the number of parameters in LwF also grows linearly with the number of tasks.
Both ResAdapt and DAN address this problem by introducing a small amount of extra parameters that adapt the source network to the target task. This is accomplished by adjusting each layer's activations in the case of ResAdapt, or their parameters directly in the case of DAN.
Although both methods can share large blocks across tasks, none try to adjust the model complexity to the target task.
As shown in Table~\ref{tab:decathlon}, \textsc{NetTailor} outperforms ResAdapt by $1.57\%$ across 10 datasets and achieves 332 points higher in the decathlon score.
More importantly, \textsc{NetTailor} only uses $3.67\!\times\!10^6$ parameters ($43\%$ fewer than ResAdapt) and $0.61\!\times\!10^9$ FLOPs ($36\%$ fewer than ResAdapt) per task.

%% file: table-skips.tex
\begin{table}[t!]
\centering
\rowcolors{2}{pastelblue}{white}
\resizebox{0.9\linewidth}{!}{
\begin{tabular}{rrcccc}
    \toprule
    && Accuracy & \# Parameters & \# FLOPS & \# Blocks \\ 
    \midrule
    \cellcolor{white} 
    & Fine-tuning & 96.59\,\% & 21.29\,M & 3.58\,B & 18 \\
    & 1-Skip & 96.60\,\% & 5.13\,M & 1.79\,B & 9 \\
    \cellcolor{white} 
    & 3-Skip & 96.53\,\% & 5.65\,M & 1.94\,B & 9 \\
    & 5-Skip & 96.79\,\% & 6.14\,M & 1.62\,B & 7 \\
    \parbox[t]{2mm}{\multirow{-5}{*}{\rotatebox[origin=c]{90}{\bf SVHN}}}
    \cellcolor{white} 
    & Dense-Skip & 96.13\,\% & \bf 4.01\,M & \bf 1.22\,B & \bf 5 \\
    \midrule
    &Fine-tuning & 95.84\,\% & 21.29\,M & 3.58\,B & 18 \\
    \cellcolor{white}
    &1-Skip      & 96.05\,\% & 8.06\,M & 2.66\,B & 13 \\
    &3-Skip      & 95.51\,\% & \bf 6.07\,M & \bf 1.85\,B & \bf 9 \\
    \cellcolor{white}
    &5-Skip      & 95.56\,\% & 7.12\,M & 2.45\,B & 9 \\
    \parbox[t]{2mm}{\multirow{-5}{*}{\rotatebox[origin=c]{90}{\bf Flowers}}}
    &Dense-Skip & 95.58\,\% & 6.45\,M & 2.29\,B & 11 \\
    \midrule
    \cellcolor{white}
    & Fine-tuning & 82.82\,\% & 21.29\,M & 3.58\,B & 18 \\
    & 1-Skip     & 82.42\,\% & \bf 12.81\,M & 3.13\,B & 15 \\
    \cellcolor{white}
    & 3-Skip     & 82.33\,\% & 13.54\,M & 2.98\,B & 14 \\
    & 5-Skip     & 82.56\,\% & 12.94\,M & \bf 2.89\,B & \bf 13 \\
    \cellcolor{white}
    \parbox[t]{2mm}{\multirow{-5}{*}{\rotatebox[origin=c]{90}{\bf VOC}}}
    & Dense-Skip & 82.56\,\% & 14.85\,M & 3.30\,B & 15 \\
    \bottomrule
\end{tabular}
}
\caption{Effect of initial student architecture.}
\label{tab:skips}
\end{table}

%% file: table-piggyback.tex
\begin{table*}[t!]
\centering
\rowcolors{2}{pastelblue}{white}
\resizebox{\linewidth}{!}{
\begin{tabular}{r|ccc|ccc|ccc|ccc|ccc|ccc}
\toprule
\rowcolor{white}
& \multicolumn{3}{c|}{CUB~\cite{cub}} & \multicolumn{3}{c|}{Cars~\cite{cars}} & \multicolumn{3}{c|}{Flowers~\cite{flowers}} & \multicolumn{3}{c|}{WikiArt~\cite{wikiart}} & \multicolumn{3}{c|}{Sketch~\cite{sketch}} &
Avg & Avg & Avg \\ 
& Acc & Params & FLOPs & Acc & Params & FLOPs & Acc & Params & FLOPs & Acc & Params & FLOPs & Acc & Params & FLOPs & Acc & Params & FLOPs \\ \midrule
Feature~\cite{mallya2018piggyback}   
& 70.03 & 23.9 & 4.11 & 52.80 & 23.9 & 4.11 & 85.99 & 23.9 & 24.0 & 55.60 & 23.9 & 4.11 & 50.86 & 23.9 & 4.11 & 63.05 & 23.9 & 4.11 \\
%
PackNet  $\rightarrow{}$ \cite{mallya2017packnet}  
& 80.31 & 23.9 & 4.11 & 86.11 & 23.9 & 4.11 & 93.04 & 23.9 & 4.11 & 69.40 & 23.9 & 4.11 & 76.17 & 23.9 & 4.11 & 81.01 & 23.9 & 4.11 \\
PackNet $\leftarrow{}$ \cite{mallya2017packnet}  
& 71.38 & 23.9 & 4.11 & 80.01 & 23.9 & 4.11 & 90.55 & 23.9 & 4.11 & 70.31 & 23.9 & 4.11 & 78.70 & 23.9 & 4.11 & 78.19 & 23.9 & 4.11 \\
Piggyback \cite{mallya2018piggyback} 
& 81.59 & 24.3 & 4.11 & 89.62 & 24.3 & 4.11 & 94.77 & 24.3 & 4.11 & 71.33 & 24.3 & 4.11 & 79.91 & 24.3 & 4.11 & 83.44 & 24.3 & 4.11 \\
\midrule
\textsc{NetTailor}
& \bf 82.52 & \bf 13.7 & \bf 3.31 & \bf 90.56 & \bf 12.9 & \bf 3.31 & \bf 95.79 & \bf 8.5 & \bf 2.37 & \bf 72.98 & \bf 15.4 & \bf 3.55 & \bf 80.48 & \bf 15.1 & \bf 3.44 & \bf 84.47 & \bf 13.1 & \bf 3.20 \\ \bottomrule
\end{tabular}}
\caption{Accuracy and model complexity for prior transfer learning methods in five datasets.
PackNet performance is sensitive to the order in which datasets are presented. $\rightarrow{}$ indicates the following order: CUB, Cars, Flowers, WikiArt and Sketch. $\leftarrow{}$ indicates reversed order.}
\label{tab:piggyback}
\end{table*}

%% file: table-decathlon.tex
\begin{table*}[t!]
\rowcolors{2}{white}{pastelblue}
\resizebox{\linewidth}{!}{
\begin{tabular}{r|cccccccccc|cccc}
\toprule
& ImNet~\cite{imagenet} & Airc~\cite{aircraft} & C100~\cite{cifar} & DPed~\cite{dped} & DTD~\cite{dtd} & GTSR~\cite{gtsr} & Flwr~\cite{flowers} & Oglt~\cite{omniglot} & SVHN~\cite{svhn} & UCF~\cite{ucf} & Mean & Score & Avg Params & Avg FLOPS \\ \midrule
LwF~\cite{li2017LwF,rebuffi2017learning}  & 59.87 & 61.15 & \bf 82.23 & 92.34 & 58.83 & 97.57 & 83.05 & 88.08 & 96.10 & 50.04 & 76.93 & 2515 & 5.86 & 0.87 \\
Piggyback~\cite{mallya2018piggyback} & 57.69 & 65.29 & 79.87 & \bf 96.99 & 57.45 & 97.27 & 79.09 & 87.63 & \bf 97.24 & 47.48 & 76.60 & 2838 & 6.04 & 0.87 \\
DAN~\cite{rosenfeld2018incremental} & 57.74 & 64.12 & 80.07 &  91.30 & 56.54 & 98.46 & 86.05 & 89.67 & 96.77 &  49.38 & 77.01 & 2851 & 6.54 & 0.97 \\
ResAdapt~\cite{rebuffi2018efficient} & 60.32 & 64.21 & 81.91 & 94.73 & 58.83 & 99.38 & 84.68 & 89.21 & 96.54 & \bf 50.94 & 78.07 & 3412 & 6.44 & 0.96 \\
\midrule
\textsc{NetTailor} & \bf 61.42 & \bf 75.07 & 81.84 & 94.68 & \bf 61.28 & \bf 99.52 & \bf 86.53 & \bf 90.09 & 96.44 & 49.54 & \bf 79.64 & \bf 3744 & \bf 3.67 & \bf 0.61 \\
\rowcolor{white}
\bottomrule
\end{tabular}
}
\caption{Accuracy and model complexity of several methods on the visual decathlon challenge. }
\label{tab:decathlon}
\end{table*}

%% file: 5_conclusion.tex
In this work, we introduced a novel transfer learning approach, denoted \textsc{NetTailor}, which adapts the \textit{architecture} of a pre-trained model to a target task.
\textsc{NetTailor} uses the layers of the pre-trained CNN as universal blocks shared across tasks and combines them with small task-specific layers to generate a new network. 
Experiments have shown that \textsc{NetTailor} is capable of learning architectures of increasing complexity for increasingly harder tasks, while achieving performances similar to that of transfer techniques like fine-tuning.